\documentclass[runningheads]{llncs}
\usepackage[T1]{fontenc}
\usepackage{graphicx}
\usepackage{mathtools}
\usepackage{amssymb}
\usepackage{caption}
\usepackage[normalem]{ulem}
\usepackage{pifont}
\usepackage[table,xcdraw]{xcolor}
\usepackage{tabulary}
\usepackage{hyperref}
\newcommand{\cmark}{\ding{51}}%
\useunder{\uline}{\ul}{}
\begin{document}
\title{\texttt{SinoSynth}: A Physics-based Domain Randomization Approach for Generalizable CBCT Image Enhancement}

\titlerunning{\texttt{SinoSynth} for Generalizable CBCT Image Enhancement}

\author{Yunkui Pang \inst{1}$^{\star}$\orcidID{0000-0003-2798-337X}
\and
Yilin Liu \inst{1}\thanks{These authors contribute equally.}
\orcidID{0000-0002-2540-1295}
\and
Xu Chen\inst{2}\orcidID{0000-0002-0367-3003}
\and
Pew-Thian Yap\inst{1}\orcidID{0000-0003-1489-2102}
\and
Jun Lian\inst{1}\orcidID{0000-0002-2041-9074}
}
\authorrunning{Y. Pang et al.}
%
\institute{University of North Carolina at Chapel Hill, Chapel Hill NC 27599, USA. \\
\email{\{yunkuipa,yilinliu\}@cs.unc.edu, \{jun\_lian,ptyap\}@med.unc.edu}\\
\and
Huaqiao University, Xiamen 361021, China.
\email{chenxu31@gmail.com}}
\maketitle              
\begin{abstract}
Cone Beam Computed Tomography (CBCT) finds diverse applications in medicine. Ensuring high image quality in CBCT scans is essential for accurate diagnosis and treatment delivery. Yet, the susceptibility of CBCT images to noise and artifacts undermines both their usefulness and reliability. Existing methods typically address CBCT artifacts through image-to-image translation approaches. These methods, however, are limited by the artifact types present in the training data, which may not cover the complete spectrum of CBCT degradations stemming from variations in imaging protocols. Gathering additional data to encompass all possible scenarios can often pose a challenge. To address this, we present \texttt{SinoSynth}, a physics-based degradation model that simulates various CBCT-specific artifacts to generate a diverse set of synthetic CBCT images from high-quality CT images, \textit{without} requiring pre-aligned data. Through extensive experiments, we demonstrate that several different generative networks trained on our synthesized data achieve remarkable results on heterogeneous multi-institutional datasets, outperforming even the same networks trained on actual data. We further show that our degradation model conveniently provides an avenue to enforce anatomical constraints in conditional generative models, yielding high-quality and structure-preserving synthetic CT images\footnote{\url{https://github.com/Pangyk/SinoSynth}}. 

\end{abstract}

\section{Introduction}
CBCT provides intricate 3D imaging capabilities crucial for precise diagnosis, treatment planning, and surgical guidance across diverse medical fields. 
\cite{gupta2013cone,bayaraa2020twostage,miracle2009conebeam}. CBCT typically utilizes lower radiation doses compared to fan-beam CT and boasts greater portability and cost-effectiveness, which are key factors driving its widespread deployment in treatment rooms. However, CBCT images are susceptible to noise and artifacts \cite{nagarajappa2015artifacts}, resulting in lower image quality than CT images \cite{venkatesh2017cone}. Thus, it has been of great interest to the clinical community to enhance the image quality of CBCT.

\begin{figure}[t]
\centerline{\includegraphics[width=0.8\linewidth]{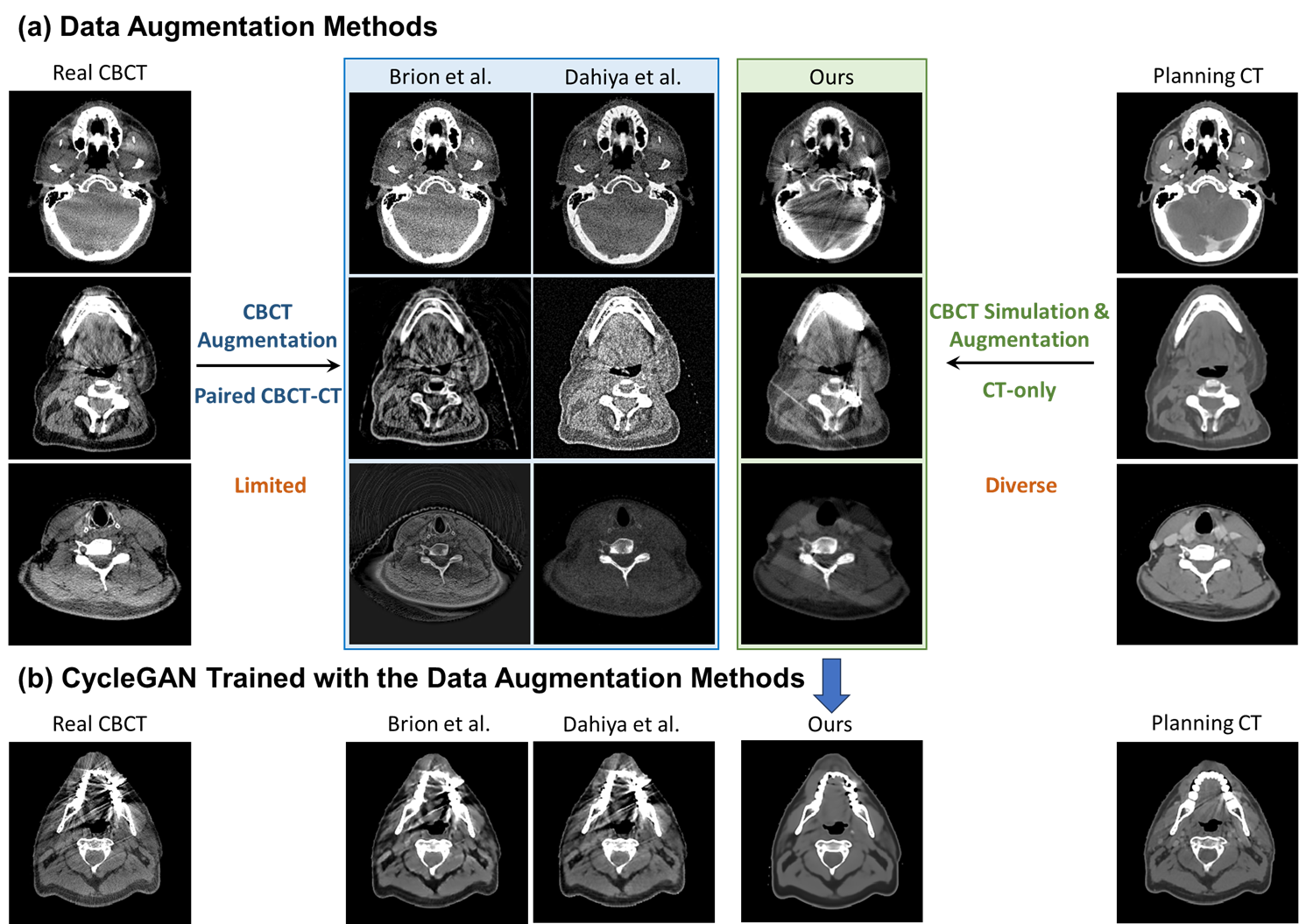}}
\caption{\textbf{(a)} Comparisons between ours and other CBCT data augmentation methods. \textbf{(b)} Our method significantly improves the denoising performance of CycleGAN without requiring pre-aligned paired datasets.}
\label{fig1}
\end{figure}


Supervised image-to-image translation methods have achieved promising performance in CBCT enhancement, under the prerequisite of sufficient paired and well-aligned CBCT-CT datasets \cite{chen2020synthetic,peng2023cbct}. Unsupervised methods utilizing generative adversarial networks (GANs) \cite{chen2021synthetic,liu2021cbct,chen2023organ}, which leverage unpaired CBCT-CT images, partially mitigate the issues. Nevertheless, their implicit learning of CBCT characteristics (e.g., artifacts) may make it challenging to distinguish between anatomical details and noise when the dataset size is small, as demonstrated in our experiments. 
Although there have been a few publicly available datasets \cite{balik2013evaluation,Hugo2016-am,Thummerer2023-tq} with CBCT-CT images for several organs, differences in imaging equipment and scanning protocols could still result in severe performance degradation of the model pre-trained on these datasets (Fig.~\ref{fig3}). This is because the quality of CBCT images is highly dependent on the specific imaging device used, leading to significant variability in appearance and data distributions across various scanning settings \cite{liang2020generalizability}. The considerable variability presents practical challenges to data collection and model generalizability in real-world clinical scenarios.

To mitigate the issues, many efforts have been devoted to enriching the training data (Fig.~\ref{fig1}). Brion et al. \cite{brion2021domain} performed intensity-based data augmentation by incorporating adjustments to brightness, contrast, and sharpness. However, such methods do not consider the physical causes of CBCT artifacts, rendering augmentation less effective. Dahiya et al. \cite{dahiya2021multitask} simulate CBCT images by directly extracting artifacts from existing CBCT images and mapping them to the corresponding CT images. As a result, their method requires accurately pre-aligned CT and CBCT images. Additionally, the types of artifacts that can be simulated remain limited to those present within the available datasets. Synthetic data generation methods have also been explored for general planar X-ray images \cite{gao2023synthetic}, but they are not tailored to 3D CBCT images.

In this work, we present \texttt{SinoSynth}, a physics-based CT-to-CBCT degradation model with adjustable parameters for extensive domain simulation. Given a CT image, our approach synthesizes a variety of CBCT artifacts controlled by random sampling parameters, such as noise level and geometry configurations. As such, an unlimited number of aligned CBCT-CT image pairs can be generated for training. Notably, our method requires only a set of CT images, and the synthesized training data are inherently aligned. To better reflect domain-specific variations, we incorporate the domain knowledge of CBCT into the degradation model in two ways: 1) we parameterize the cone-beam geometry to simulate diverse CBCT scanning configurations, and 2) translate the formation of CBCT artifacts into algorithms controlled by adjustable parameters, adhering to X-ray properties \cite{Beer1852-hu}. As some of the artifacts typically co-occur, our focus is on simulating five representative CBCT artifacts \cite{nagarajappa2015artifacts,schulze2011artefacts}. We empirically verify the effectiveness of their combinations in covering the degradation space of actual CBCT images compared to previous augmentation methods through experiments. 

We integrate \texttt{SinoSynth} into several existing synthetic CT generation frameworks, including Denoising Diffusion GAN (DDGAN) \cite{xiao2022tackling}, CycleGAN \cite{chen2021synthetic}, DRIT \cite{lee2020drit++}, ROI-aware DCLGAN \cite{chen2023organ}, and FGDM \cite{li2023zero}. Moreover, our degradation model enables structural guidance by enforcing consistency between the synthesized CT output and the conditioned simulated CBCT image. Particularly, to cope with the stochasticity in generative networks, we showcase that this strategy ensures better structural preservation than using the original conditioning scheme alone.  

To summarize, our contributions are three-fold:
\begin{enumerate}
    \item We proposed a physics-based CBCT degradation model (\texttt{SinoSynth})  that incorporates domain knowledge to simulate domain-randomized CBCT artifacts across various imaging protocols, thereby mitigating the need for extensive data collection.
    \item Compared to networks trained with previous augmentation methods or actual data, \texttt{SinoSynth}-trained networks demonstrate significantly better zero-shot generalization ability and structure-preserving ability on challenging datasets collected from multiple hospitals.
    \item Our work underscores the significance of accurate CBCT simulation modeling for generalizable CT synthesis.
\end{enumerate}

\section{Method}
\begin{figure}[t!]
\centerline{\includegraphics[width=\linewidth]{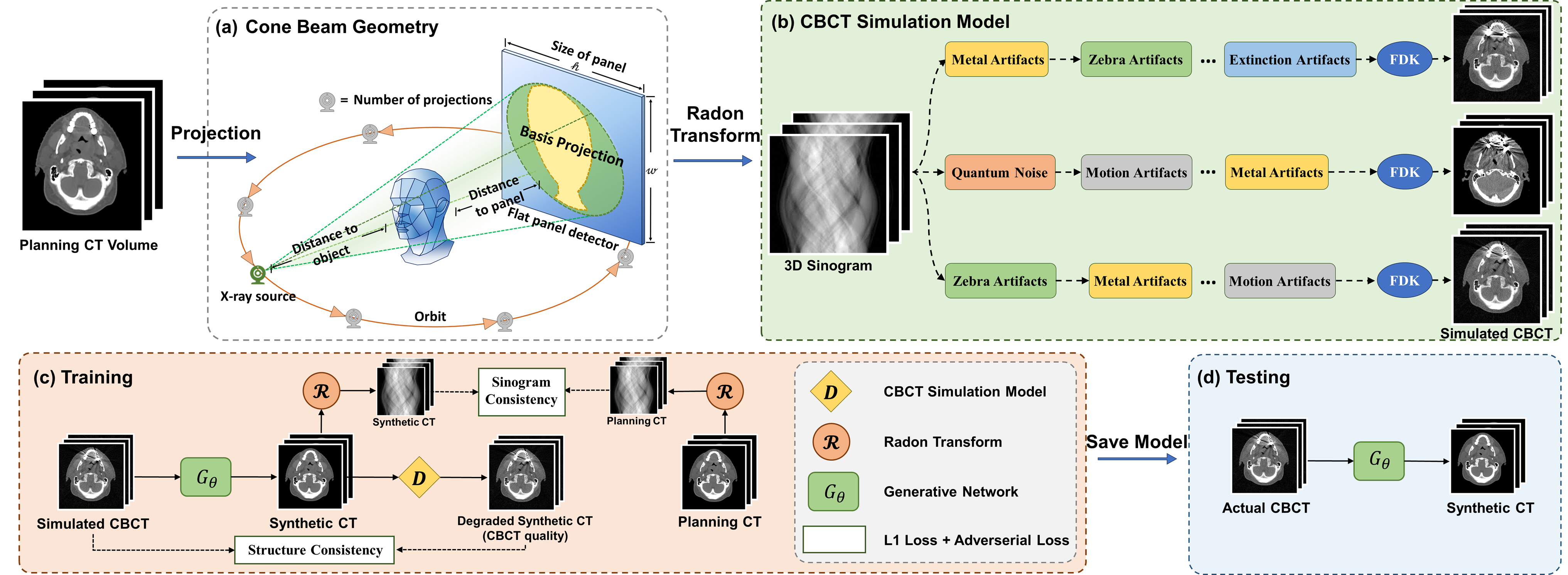}}
\caption{Overview of the proposed framework. \textbf{(a)} A planning CT volume is first transformed into the 3D sinogram. \textbf{(b)} Simulated CBCT-specific artifacts are randomly applied to the sinogram, which are then transformed back to a CBCT-quality volume using the FDK algorithm. \textbf{(c)} The simulated CBCT is fed into the generative network trained with the proposed sinogram and structure consistency constraints. \textbf{(d)} During testing, the network takes the actual CBCT as input and outputs a denoised synthetic CT image.}
\label{fig2}
\end{figure}

Fig.~\ref{fig2} illustrates the proposed pipeline. \texttt{SinoSynth} generates sCBCT images by simulating CBCT artifacts on CT images in each batch on the fly during training (Sec.~\ref{sec:simulation}). The artifacts are simulated with random occurrences and a shuffled order. Meanwhile, the degradation model is used for structural guidance (Sec.~\ref{sec:structural}). After training, the network is directly applied to actual CBCT images.

\subsection{Preliminaries}
Our CBCT simulation is performed in the sinogram domain. The CT slice $x$ represented in Hounsfield Units is first converted into the linear attenuation coefficient map $x_\mu$ via a linear transformation \cite{brown2008investigation}. \texttt{Radon} transform $\mathcal{R}$ integrated along the cone beam geometry $G$ is then applied to project $x_\mu$ onto the sinogram:
\begin{equation}
    x_s = -ln (\int_{E} I_0(E_i) \cdot \exp(-\int_{G}(x_\mu(t) \cdot \frac{m_{E_i}}{m_{E_0}})dt)dE)
    \label{eq_sinogram}
\end{equation} where the constant $I_0(E_i)$ is the intensity of the entry X-ray beam within composite energy levels $\{E_i| i \in [1, N]\} \subseteq [20,120]$ keV. We then simulate different types of CBCT artifacts and noise based on the derived sinogram $x_s$. Reconstructing from the sinogram using filtered back projection gives the corresponding CBCT image that is used for training. We implemented Eq~\ref{eq_sinogram} with the ASTRA Toolbox \cite{van2016fast} and Operator Discretization Library \footnote{\url{https://github.com/odlgroup/odl}}.

\subsection{Domain-Randomized CBCT Simulation} \label{sec:simulation}

\textbf{Scanner Effects Simulation.} The diversity among different CBCT scanners significantly impacts the varied appearance of CBCT artifacts, resulting in decreased performance of the network \cite{liang2020generalizability}. To simulate the scanner variability, we parameterize CBCT scanners with different cone beam geometries $G(d_1, d_2, n, s)$. As illustrated in Fig.~\ref{fig2} (b), $d_1 \in [10, 100]$ denotes the distance from the X-ray beam source to the patient; $d_2 \in [10, 100]$ is the distance from the patient to the detector; $n \in [64, 512]$ denotes the number of X-ray projections, and a smaller $n$ indicates a sparser view; $s \in [128\times128, 512\times512]$ denotes the detector size. Hence, we simulate the effects of CBCT scanners from different vendors by computing Eq.~\ref{eq_sinogram} with randomly parameterized $G$.


\textbf{Metal Artifacts} usually occur in the presence of metal implants. The low X-ray transmission of metallic implants and the polychromatic nature of the X-ray source result in severe beam hardening \cite{bayaraa2020twostage}, bringing scatter and streak artifacts \cite{8331163}. Thus, for simulation, we need to create metal trace regions on the sinogram $x_s$. This is done by first taking the intersection between a random mask $m_t$ obtained through cubic bézier curve \cite{han2008novel} controlled by randomly sampled points, and a mask of bone areas $m_b$ obtained through thresholding, $m_{\Omega} = m_t \cap m_b$, and converting $m_{\Omega}$ to the sinogram $m_{\Omega}^s$. Then, we update the metal trace regions of $x_s$ by filling in the linear attenuation coefficients of the metal implanted area:
\begin{equation}
    y_s = x_s + \sum_{i}^{E}m_{\Omega}^s \cdot metal_\mu(E_i),
\end{equation} where each $metal_\mu(E_i)$ corresponding to an energy level $E_i$ is computed as: $metal_\mu(E_i) = m_{m} \cdot \mu_{m}(E_i) \cdot (\rho_{m} - 1.0)$. 
The mass attenuation coefficient $\mu_{m}(E_i)$ can be obtained from \cite{GERWARD1993783}. Adjusting the density of the metal material $\rho_{m} > 1$ allows for controlling the intensity of the metal artifacts.

\textbf{Extinction Artifacts} occur when the object contains highly absorbent material, which significantly attenuates the X-ray signal, reducing it to near-zero \cite{schulze2011artefacts}. Since the attenuated regions on CBCT images are often irregular and discontinuous, we generate a random image-domain mask $m_{et}^s$ using the cubic bézier curve controlled by randomly sampled points within a local region. The sinogram $x_s$ is then updated by:
\begin{equation}
    y_s =  \mathcal{R}(\lambda_{et} \cdot m_{et} \cdot x) + \mathcal{R}((1 - m_{et}) \cdot x) 
\end{equation}
where $\mathcal{R}$ denotes \texttt{Radon} transform, $\lambda_{et} \in [0,1]$ controls the attenuation extent.

\textbf{Quantum Noise} is the main source of image deterioration in plain radiography, which arises from the natural variability in how photons reach the detector \cite{wang2008experimental}. As quantum noise predominantly impacts images acquired using low radiation doses, CBCT images are expected to exhibit a higher noise level compared to CT images. Given that quantum noise usually follows a Poisson distribution \cite{strid1980significance}, the quantum noise on sinogram $x_s$ can be simulated as:
\begin{equation}
    y_s = \frac{(\alpha \cdot x_s)^k \cdot e^{-\alpha \cdot x_s}}{\alpha \cdot k!}
\end{equation}
where $\alpha$ controls the noise level, and $k$ is the number of photon occurrences during a time interval. 

\textbf{Zebra artifacts} appear as alternating bright and dark stripes in CBCT images due to helical interpolation \cite{nagarajappa2015artifacts}. We simulate them by creating a binary mask $m_z$ with $n\sim\mathcal{U}(0, N)$ stripes. The width of each stripe is randomized within the range of $[1, 10]$ pixels. We then simulate stripes of different directions by applying rotation to the mask $m_z = \texttt{Rotate}(m_{z}, \theta)$, $\theta \in [0, \pi)$. The zebra artifact on the sinogram $x_s$ is simulated as:
\begin{equation}
    y_s = \mathcal{R}(m_z \cdot v \cdot x) + \mathcal{R}((1 - m_z) \cdot x), v\sim \mathcal{U}(0, \mathbf{I})
\end{equation} 

\textbf{Motion Artifacts} are caused by the motion of the patient that appears as unsharpness in the reconstructed image \cite{nagarajappa2015artifacts}. The process can be viewed as an image $x$ disturbed by small deformations and displacements $\mathcal{D}$ during the imaging process, along with a blurring effect simulated by the Gaussian filter:
\begin{equation}
    y_s = \mathcal{R}(\mathcal{G}(p \cdot x + (1 - p) \cdot \mathcal{DM}(\sigma_d, r, z)(x))),
\end{equation} where $\mathcal{G}$ is the Gaussian blur filter. $\mathcal{DM}(\sigma_d, r, z)$ represents the deformation and displacement caused by the motion, controlled by the deformation degree $\sigma_D \in [0.5, 1.5]$, the rotation angle $r \in [0, \frac{\pi}{60}]$ and the resizing factor $z \in [0.9, 1.1]$. $p$ is a factor balancing $x$ and its deformed result $\mathcal{D}(\cdot)(x)$.

\subsection{Structural Guidance for Generative networks } \label{sec:structural}
As our CBCT simulation model $\mathcal{D}$ is differentiable, we propose simple anatomical structure constraints to regularize the generation of sCT in both the sinogram and image domains. For a simulated CBCT image $y$ and the network output $G_\theta(y)$, the denoised output should be converted back to $y$ via the degradation model. Meanwhile, the output should be consistent with the reference CT in the sinogram domain after being applied with a mask $M$ that removes the metal-affected region. Hence, the generative network $G_\theta$ is guided by the following losses to generate anatomically consistent output:
\begin{equation}
    \mathcal{L}_{\text{struc}} = \mathbb{E}_{y}\left[\|\mathcal{D}(G_\theta(y)) - y\|_1\right].
\end{equation}
\begin{equation}
    \mathcal{L}_{\text{sino}} = \mathbb{E}_{x, y}\left[\|M \odot (\mathcal{R}(G_\theta(y)) - \mathcal{R}(x))\|_1\right].
\end{equation}
Since we applied our method to existing frameworks, all models were trained with their original losses (e.g., GAN losses) alongside the proposed structure-preserving losses.

\begin{figure}[htbp]
\centering
\begin{minipage}{\linewidth}
\centering
  \scriptsize
\captionof{table}{Quantitative results comparing networks trained with actual CBCT-CT data and our simulated CBCT-CT data. Our method significantly improves the denoising and CBCT image enhancement performance of the networks.}
\begin{tabular}{l|cccc|cccc}
\hline
Data             & \multicolumn{4}{c|}{\textbf{Actual CBCT-CT data}}                                                             & \multicolumn{4}{c}{Our \textbf{Simulated CBCT-CT data}}                                                     \\ \hline
Methods          & \multicolumn{1}{l}{Supervised \cite{chen2020synthetic}} & \multicolumn{1}{c}{DCLGAN \cite{chen2023organ}} & \multicolumn{1}{c}{DRIT \cite{lee2020drit++}}  & FGDM \cite{li2023zero}  & \multicolumn{1}{c}{Supervised}  & \multicolumn{1}{l}{DCLGAN\phantom{s}} & \multicolumn{1}{c}{DRIT}  & FGDM           \\ \hline
\multicolumn{1}{l|}{PSNR $\uparrow$}  & \multicolumn{1}{c}{24.82}      & \multicolumn{1}{c}{23.85}  & \multicolumn{1}{c}{22.54\phantom{s}} & 23.36 & \multicolumn{1}{c}{{\ul 25.44}} & \multicolumn{1}{c}{24.34}  & \multicolumn{1}{c}{23.14} & \textbf{25.61} \\ \hline
\multicolumn{1}{l|}{SSIM $\uparrow$}  & \multicolumn{1}{c}{0.821}      & \multicolumn{1}{c}{0.784}  & \multicolumn{1}{c}{0.758\phantom{s}} & 0.819 & \multicolumn{1}{c}{{\ul 0.829}} & \multicolumn{1}{c}{0.801}  & \multicolumn{1}{c}{0.773} & \textbf{0.838} \\ \hline
\multicolumn{1}{l|}{MAE $\downarrow$} & \multicolumn{1}{c}{32.73}      & \multicolumn{1}{c}{35.04}  & \multicolumn{1}{c}{80.97\phantom{s}} & 36.42 & \multicolumn{1}{c}{{\ul 26.32}} & \multicolumn{1}{c}{28.34}  & \multicolumn{1}{c}{31.24} & \textbf{22.50} \\ \hline
\end{tabular}
  \label{tab1}
\end{minipage}
\par\medskip 

\begin{minipage}{\linewidth}
  \centering
\centerline{\includegraphics[width=\linewidth]{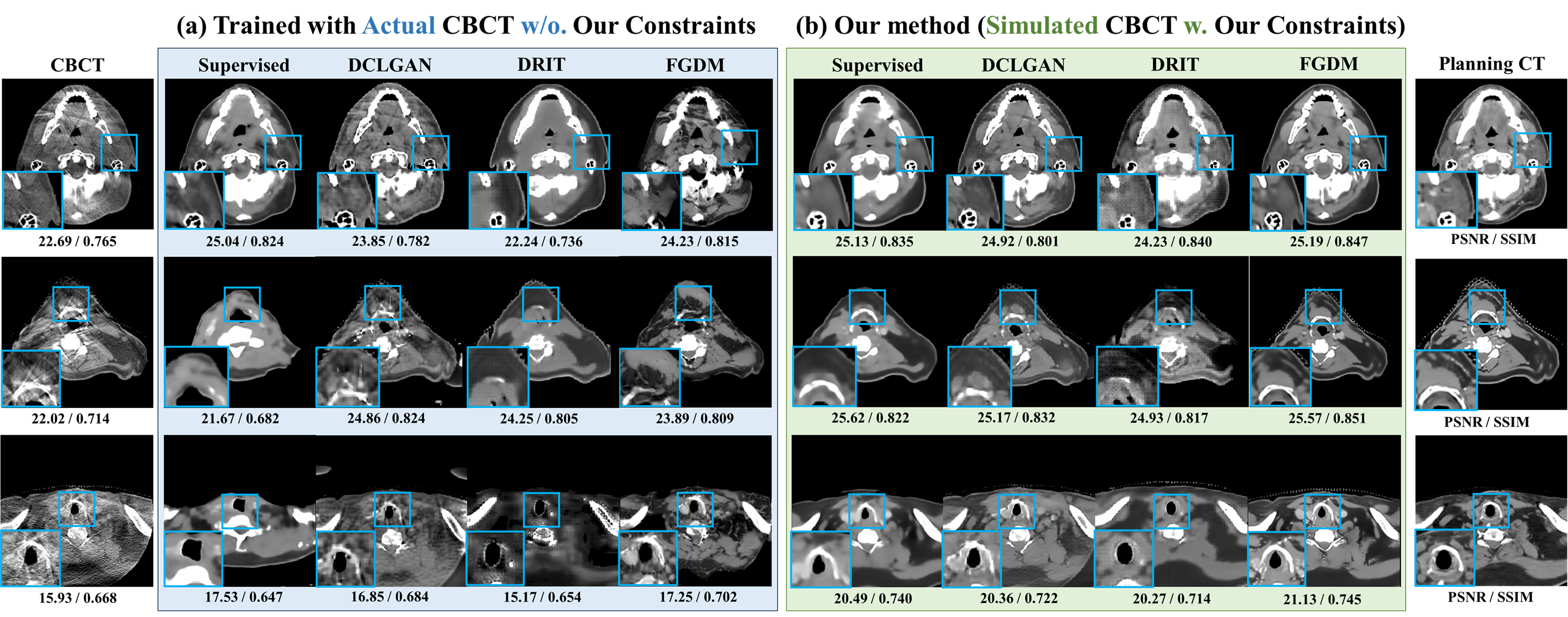}}
\caption{Qualitative comparisons between networks trained with actual CBCT-CT data and our simulated CBCT-CT data.}
\label{fig3}
\end{minipage}
\end{figure}


\begin{figure}[htbp]
\centering
\begin{minipage}{\linewidth}
\centering
\scriptsize
\captionof{table}{Quantitative evaluations of different data augmentation methods applied to CycleGAN. Our method significantly improves the denoising and CBCT image enhancement performance of the base model.}
\begin{tabulary}{0.9\linewidth}{l|cccc}
\hline
Methods & \phantom{s}w/o. Aug\phantom{s} & \phantom{s}Brion et al.\phantom{s} & \phantom{s}Dahiya et al.\phantom{s} & \phantom{s}Ours\phantom{s}           \\ \hline
PSNR $\uparrow$   & 22.51    & 23.67        & 23.84   & \textbf{25.08} \\ 
SSIM $\uparrow$    & 0.733    & 0.752        & 0.769   & \textbf{0.823} \\ 
MAE $\downarrow$     & 60.07    & 51.52        & 50.18   & \textbf{32.67} \\ \hline
\end{tabulary}
\label{tab2}
\end{minipage}


\begin{minipage}{\linewidth}
  \centering
  \includegraphics[width=0.8\linewidth]{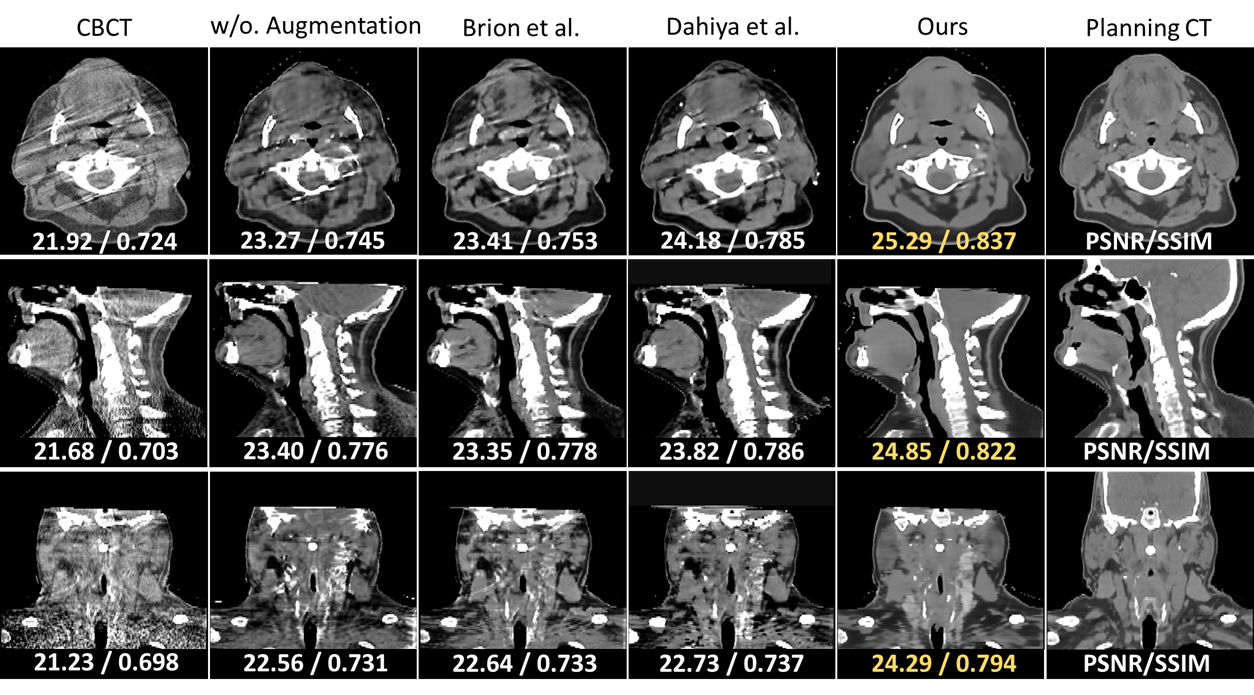}
  \caption{Qualitative evaluations of different data augmentation methods.}
  \label{fig4}
\end{minipage}
\end{figure}

\section{Experiments} \label{sec:exp}
\subsection{Datasets and Pre-processing} We curated a Head and Neck CBCT dataset with 250 patients from five hospitals in Europe and the US with mAs used in scanning (10, 12, 20, and 50). 160 randomly sampled patient data are used for training, 20 for validation, and 70 for evaluation. All the data are resampled to a voxel size of $1 \times 1 \times 2 \textit{mm}^3$, with an image size of $256 \times 256$ for each slice. We utilize Peak-to-Signal Ratio (PSNR \cite{5596999}), Structure Similarity (SSIM \cite{5596999}), and Mean Absolute Error (MAE) of the HU values for evaluating all compared methods. Our method simulates CBCT images on the fly during training instead of pre-generating them. Each model is trained for 200 epochs.

\subsection{Comparisons with actual CBCT-CT data}
We compare the established image-to-image translation models trained on our simulated data to the same models trained on the actual CBCT-CT data. The base models include supervised U-Net \cite{chen2020synthetic}, ROI-aware DCLGAN \cite{chen2023organ}, DRIT \cite{lee2020drit++}, and FGDM \cite{li2023zero} based on diffusion probabilistic model \cite{xiao2022tackling}. 
Table~\ref{tab1} and Fig.~\ref{fig4} demonstrate that our method significantly enhances generalization across various artifacts, leading to the superior performance of CBCT noise reduction and soft tissue enhancement. Our method also improves the inpainting performance of the shoulder region. 

\subsection{Comparisons with other augmentation methods}
We compare \texttt{SinoSynth} with existing augmentation methods \cite{dahiya2021multitask,brion2021domain}. CycleGAN \cite{chen2021synthetic} is employed as the base model. As shown in Fig.~\ref{fig3}, and Table~\ref{tab2}, \texttt{SinoSynth} outperforms the existing augmentation methods. This is because the existing methods inadequately amplify the diversity of CBCT-specific artifacts, rendering the network susceptible to out-of-distribution CBCT artifacts, as depicted in Fig.~\ref{fig1}. Our approach addresses this limitation by explicitly simulating a wide range of CBCT artifacts, thereby enhancing the performance of the CycleGAN.

\begin{table}[htbp]
  \centering
\scriptsize
\caption{Ablation study on the influences of the simulated CBCT artifacts. DDGAN \cite{xiao2022tackling} is employed as the base model for evaluation. \textit{Si-Cons.}: Sinogram Consistency Constraint. \textit{St-Cons.}: Structure Consistency Constraint.}
\vspace{3pt}
\begin{tabular}{ccccc|cc|ccc}
\hline
Metal                 & Quantum               & Extinction            & Zebra                 & Motion                & \multicolumn{1}{l}{Si-Cons.} & \multicolumn{1}{l|}{St-Cons.} & PSNR $\uparrow$ & SSIM $\uparrow$ & MAE $\downarrow$ \\ \hline
\cmark &                       &                       &                       &                       &                                           &                                            & 23.01           & 0.733           & 32.63            \\ 
                      & \cmark &                       &                       &                       &                                           &                                            & 23.62           & 0.732           & 31.24            \\ 
                      &                       & \cmark &                       &                       &                                           &                                            & 22.24           & 0.722           & 32.36            \\ 
                      &                       &                       & \cmark &                       &                                           &                                            & 23.16           & 0.741           & 32.45            \\ 
                      &                       &                       &                       & \cmark &                                           &                                            & 22.87           & 0.716           & 33.57            \\ \hline
\cmark & \cmark & \cmark & \cmark & \cmark & \cmark                     &                                            & 23.28           & 0.792           & 29.82            \\ 
\cmark & \cmark & \cmark & \cmark & \cmark &                                           & \cmark                      & 24.56           & 0.774           & 25.51            \\ 
\cmark & \cmark & \cmark & \cmark & \cmark & \cmark                     & \cmark                      & \textbf{25.63}  & \textbf{0.841}  & \textbf{21.44}   \\ \hline
\end{tabular}
\label{tab3}
\end{table}

\subsection{Ablation studies} Quantitative results in Table.~\ref{tab3} reveal two key insights. Firstly, for ablating the artifact types, PSNR/SSIM scores imply the occurrence frequency of the CBCT artifacts present in the test data. Secondly, for ablating the loss function design choices, both constraints contribute to CBCT image enhancement. As shown qualitatively in the appendix, the structure consistency constraint plays a crucial role in synthesizing structurally consistent images. The sinogram consistency constraint aids both CT noise reduction and accurate structure reconstruction.

\section{Conclusion}
In this work, we introduce a physics-based domain randomization approach to address the inherent challenges associated with generating corresponding CT images from CBCT scans, including susceptibility to artifacts and limited generalizability.  Our innovative approach involves synthesizing CBCT images with realistic artifacts, enabling us to overcome these obstacles. Through extensive experiments, we demonstrate that deep generative networks trained on our synthetic CBCT images outperform those trained on actual data.  This suggests a promising avenue for leveraging simulated CBCT data to train deep networks on larger-scale CT-only datasets, which are more readily accessible online. Our work not only improves the reliability of CBCT in clinical settings but also lays the groundwork for future advancements in other medical imaging modalities. 

\begin{credits}
\subsubsection{\ackname} This work was supported by the National Institutes of Health under grants R01CA206100 and R01EB035160 and UNC Lineberger Developmental Award 29242. Xu Chen was supported by NSFC grant 62276105, Natural Science Foundation of Xiamen China 3502Z20227193, Natural Science Foundation of Fujian Province 2023J01136, and Scientific Research Funds of Huaqiao University 20221XD029.

\subsubsection{\discintname}
The authors have no competing interests to declare that are relevant to the content of this article.
\end{credits}
 
{
    \small
    \bibliographystyle{splncs04}
    \bibliography{Paper-1155.bib}
}

\end{document}